\documentclass[letterpaper, 10 pt, conference]{ieeeconf}  

\IEEEoverridecommandlockouts                              

\usepackage{graphicx}
\usepackage{mathtools}
\usepackage{amssymb,amsfonts}

\usepackage[dvipsnames]{xcolor}
\usepackage[pdfusetitle]{hyperref}
\usepackage{booktabs}
\usepackage{balance}
\usepackage{siunitx}
\hypersetup{pdfauthor=author}

\usepackage{bm}
\usepackage{mathrsfs}

\newcommand{\RR}{\mathbb{R}}

\newcommand{\calA}{\mathcal{A}}

\newcommand{\calC}{\mathcal{C}}

\newcommand{\calL}{\mathcal{L}}
\newcommand{\calM}{\mathcal{M}}
\newcommand{\calQ}{\mathcal{Q}}

\newcommand{\suchthat}{\mathrm{s.t.}\ }

\newcommand{\defeq}{\stackrel{\mathclap{\tiny\mathrm{def}}}{=}}

\newcommand{\bfm}[1]{\mathbf{#1}}

\DeclareMathOperator*{\argmin}{argmin}

\pdfoutput=1

\renewcommand{\leq}{\leqslant}
\renewcommand{\geq}{\geqslant}

\newcommand{\proxnlp}{\texttt{proxnlp}}

\title{\LARGE \bf
    ProxNLP: a primal-dual augmented Lagrangian solver\\for nonlinear programming in Robotics and beyond
}

\author{%
    Wilson Jallet\textsuperscript{a,b,*},~%
    Antoine Bambade\textsuperscript{b,c},~%
    Nicolas Mansard\textsuperscript{a}~and~%
    Justin Carpentier\textsuperscript{b}
    \thanks{\textsuperscript{a}~LAAS-CNRS, 7 Avenue du Colonel Roche, F-31400 Toulouse, France}%
    \thanks{\textsuperscript{b}~Inria, Département d’informatique de l’ENS, \'Ecole normale supérieure, CNRS, PSL Research University, Paris, France}%
    \thanks{\textsuperscript{c}~ENPC, France,~\textsuperscript{*}{corresponding author}:
    \href{mailto:wjallet@laas.fr}{wjallet@laas.fr}}%
}

\begin{document}
\bstctlcite{IEEEexample:BSTcontrol}

\maketitle
\thispagestyle{empty}
\pagestyle{empty}

\begin{abstract}
Mathematical optimization is the workhorse behind several aspects of modern robotics and control. In these applications, the focus is on constrained optimization, and the ability to work on manifolds (such as the classical matrix Lie groups), along with a specific requirement for robustness and speed.
In recent years, augmented Lagrangian methods have seen a resurgence due to their robustness and flexibility, their connections to (inexact) proximal-point methods, and their interoperability with Newton or semismooth Newton methods.
In the sequel, we present primal-dual augmented Lagrangian method for inequality-constrained problems on manifolds, which we introduced in our recent work, as well as an efficient C++ implementation suitable for use in robotics applications and beyond.

\textit{Paper Type} -- Recent Work~\cite{jalletConstrainedDifferentialDynamic2022} under review. (Extended with open-sourced implementation)

\end{abstract}

\section{Introduction}

The setting of optimization on manifolds is of great interest in the field of robotics, where \textit{generalized coordinates} are naturally represented using Lie groups~\cite{murrayMathematicalIntroductionRobotic2017}.
Further, solvers for robotics need to account for physical constraints such as joint angle and torque limits as well as friction cones, but also for task-based constraints which could replace penalties or costs.
Problems such as trajectory optimization or inverse dynamics with various task and physical constraints are naturally expressed as nonlinear programs (NLP).

A generic nonlinear program on a manifold $\calM$ reads as follows:
\begin{equation}\label{eq:generic_NLP}
	\begin{aligned}
		&\min_{x\in \calM} f(x)\\
		&\suchthat c(x) \in \calC
	\end{aligned}
\end{equation}
where $c\colon \calM \to \RR^m$ is a (potentially nonlinear) mapping and $\calC \subset \RR^m$ is the constraint set.

\paragraph*{Equality and inequality-constrained case}
We consider the following generic problem, which captures most problems in nonlinear optimization, including in robotics:
\begin{equation}\label{eq:generic_NLP2}
		\min_{x\in \calM} f(x)  \
		\suchthat \ g(x) = 0, \ h(x) \leq 0.
\end{equation}
Most problems of interest in robotics can be expressed this way: dynamics as equality constraints, target reaching, obstacle avoidance and friction cones as inequality constraints.

Our proposed approach is based on the augmented Lagrangian method of multipliers \cite{hestenesMultiplierGradientMethods1969,powellAlgorithmsNonlinearConstraints1978,rockafellarMultiplierMethodHestenes1973}, and its primal-dual variant \cite{gillPrimaldualAugmentedLagrangian2012}. It was first introduced in\footnote{Paper under review.} \cite{jalletConstrainedDifferentialDynamic2022} where we provide an application to constrained numerical optimal control with a novel variant of the differential dynamic programming (DDP) algorithm. The applicability of augmented Lagrangians to equality-constrained DDP was recently investigated in the robotics literature~\cite{howellALTROFastSolver2019,kazdadiEqualityConstrainedDifferential2021}, with an extension to multiple-shooting implicit dynamics in~\cite{jalletImplicitDifferentialDynamic2022}.

Overall, our key contribution is an open-source C++ solver for constrained optimization on manifolds for robotics, named  \proxnlp\footnote{\url{https://github.com/Simple-Robotics/proxnlp}.}, which relies on a novel variant of the augmented Lagrangian method.

\section{Methodology}

\subsection{Generalized primal-dual augmented Lagrangians}

This approach was first introduced for equality-constrained problems in \cite{gillPrimaldualAugmentedLagrangian2012}. We recently provided an extension to inequality-constrained problems in~\cite{jalletConstrainedDifferentialDynamic2022} with an application to constrained DDP. This method was further applied to convex QPs in Bambade et al.~\cite{bambadeProxQP}.

The classical (Hestenes-Powell-Rockafellar) augmented Lagrangian function for the problem \eqref{eq:generic_NLP2} reads:
\begin{equation}
	\label{eq:AugmentedLagrangian}
	\calL_\mu(x; y_e, z_e) = f(x) + \tfrac{1}{2\mu}\|g(x) + \mu y_{e}\|_2^2
	+ \tfrac{1}{2\mu}\|[h(x) + \mu z_e]_+\|
	.
\end{equation}
Augmented Lagrangians are known to be \textit{exact} penalty functions for constrained optimization, as in their exists an estimate $(\bar{y}, \bar z)$ and penalty parameter $\bar{\mu} > 0$ such that a minimizer $x^*$ of $\calL_{\bar{\mu}}(\cdot; \bar{y}, \bar{z})$ is a solution of \eqref{eq:generic_NLP2}.

\paragraph*{Method of multipliers}
The \textit{method of multipliers} algorithm consists in iteratively minimizing the augmented Lagrangian and taking a (projected) dual ascent step in the multipliers:
\begin{equation}\label{eq:MethodMult}
\begin{aligned}
	x^{l+1} &= \argmin_x \calL_\mu(x; y^l, z^l),\\
	y^{l+1} &= y_e + \tfrac{1}{\mu}g(x^{l+1}) \\
	z^{l+1} &= [z_e + \tfrac{1}{\mu}h(x^{l+1})]_+
\end{aligned}
\end{equation}
This process can also be seen as a proximal-point algorithm for the dual problem to the initial NLP~\cite{rockafellarAugmentedLagrangiansApplications1976}.

\paragraph*{Primal-dual function}
The \textit{primal-dual} augmented Lagrangian (pdAL) adds a penalty term for dual variables:
\begin{equation}\label{eq:pdAL_function}
\begin{split}
	&\calM_\mu(x, y, z; y_e, z_e) \defeq \calL_\mu(x; y_e, z_e)
	\\ &{\color{Orange}
	+ \tfrac{1}{2\mu}\| g(x) + \mu (y_e - y) \|_2^2
	+ \tfrac{1}{2\mu} \| [h(x) + \mu z_e]_+ - \mu z\|_2^2
}
\end{split}
\end{equation}
Any stationary point $(x^*, y^*, z^*)$ of $\calM_\mu(\cdot; y_e, z_e)$ will satisfy the KKT conditions of the iteration \eqref{eq:MethodMult} where $y^{l+1} = y^*$.

\subsection{The primal-dual Newton step}
At a nominal point $(x^k, y^k, z^k)$, the primal-dual \mbox{(quasi-)Newton} step for \eqref{eq:pdAL_function} is given by a system of equations equivalent to
\begin{equation}
	\begin{bmatrix}
	H	& g_x^\top & Ph_x^\top \\
	g_x	& -\mu I & 0\\
	Ph_x& 0		 & -\mu P
	\end{bmatrix}
	\begin{bmatrix} \delta x \\ \delta y \\ \delta z
	\end{bmatrix}
	= -\begin{bmatrix}
		\nabla\calL(x^k, y^k)  \\
		g(x^k) + \mu (y^k - y_e) \\
		[h(x^k) + \mu z^k]_+ - \mu z_e
	\end{bmatrix}
\end{equation}
where $H$ approximates the Lagrangian Hessian $\nabla^2\calL$, and $P$ is a selection matrix for rows of the matrices corresponding to the \textit{active set of constraints} $\calA(x^k)$, defined as follows:
\begin{equation}
	i \in \calA(x) \Leftrightarrow
	(h(x) + \mu z_e)_i \geq 0.
\end{equation}
As shown in \cite{gillPrimaldualAugmentedLagrangian2012}, this primal-dual step $(\delta x, \delta y, \delta z)$ is a descent direction for the pdAL function \eqref{eq:pdAL_function}.

\section{Experiments}

For solving generic NLPs, we recently implemented our method in a C++ software library named \proxnlp.
We use Eigen as our linear algebra backend~\cite{eigenweb}.
We provide an interface for rigid-body dynamics and classical matrix Lie groups (e.g. $\mathrm{SE}(3)$) using the Pinocchio~\cite{carpentierPinocchioLibraryFast2019} library which also provides derivatives~\cite{carpentierAnalyticalDerivativesRigid2018}. We also provide Python bindings.
Another C++ software package specifically dedicated to solving control problems using our variant of the DDP algorithm (exploiting the problem structure for increased efficiency) detailed in~\cite{jalletConstrainedDifferentialDynamic2022} is currently under development.

\paragraph{Simple barycenter on manifold}

\proxnlp~is able to operate on manifolds. We can quickly compute the barycenter of a few points: in this case our method reduces to Newton/Gauss-Newton iterations. See fig.~\ref{fig:se2_barycenter} below.
\begin{figure}[ht!]
	\centering
	\includegraphics[width=\linewidth]{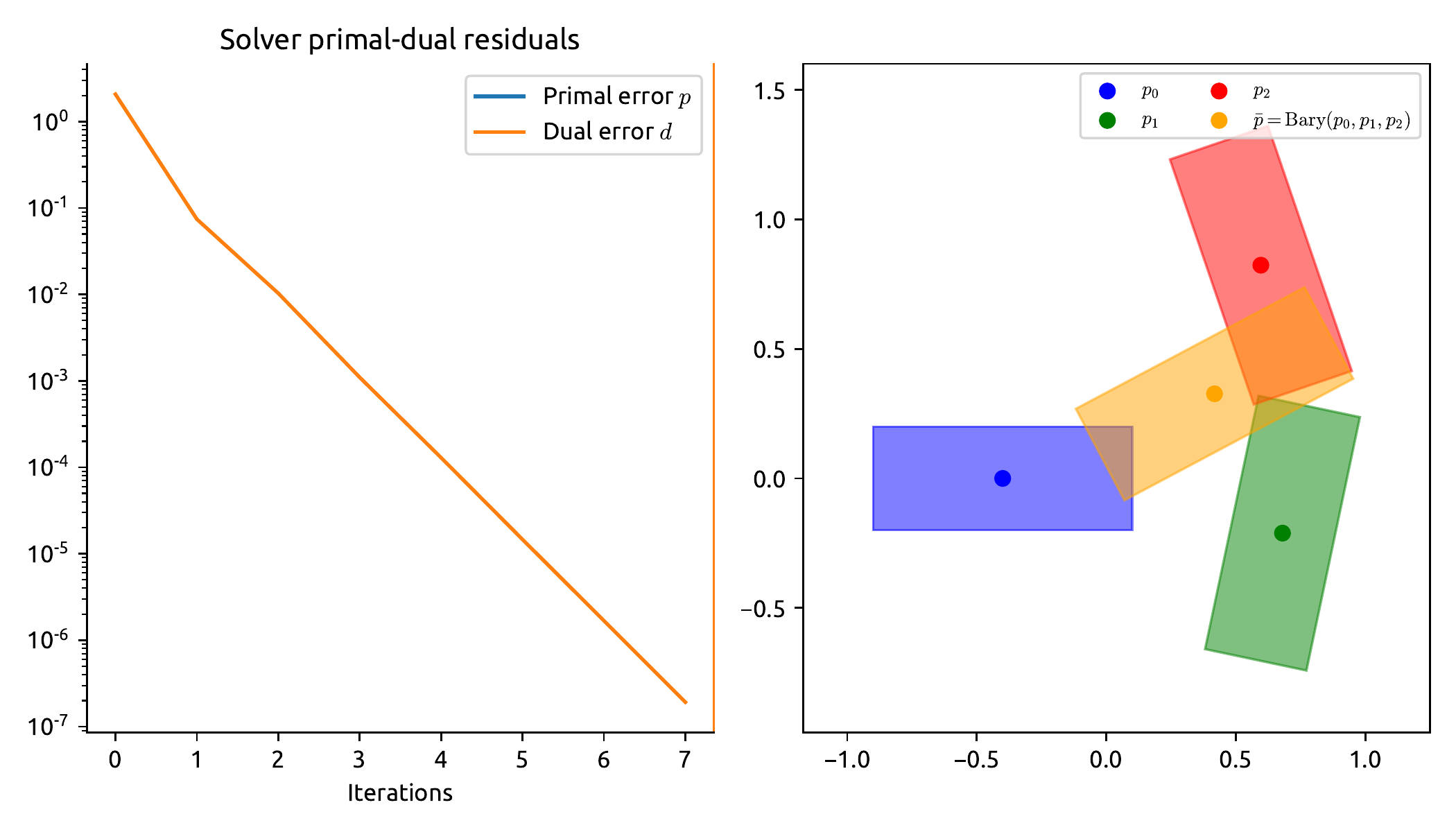}
	\caption{Barycenter of three 2D poses in the $\mathrm{SE}(2)$ Lie group.}
	\label{fig:se2_barycenter}
\end{figure}

\paragraph{Double-pendulum}

\begin{figure}[ht!]
	\centering
	\includegraphics[width=\linewidth]{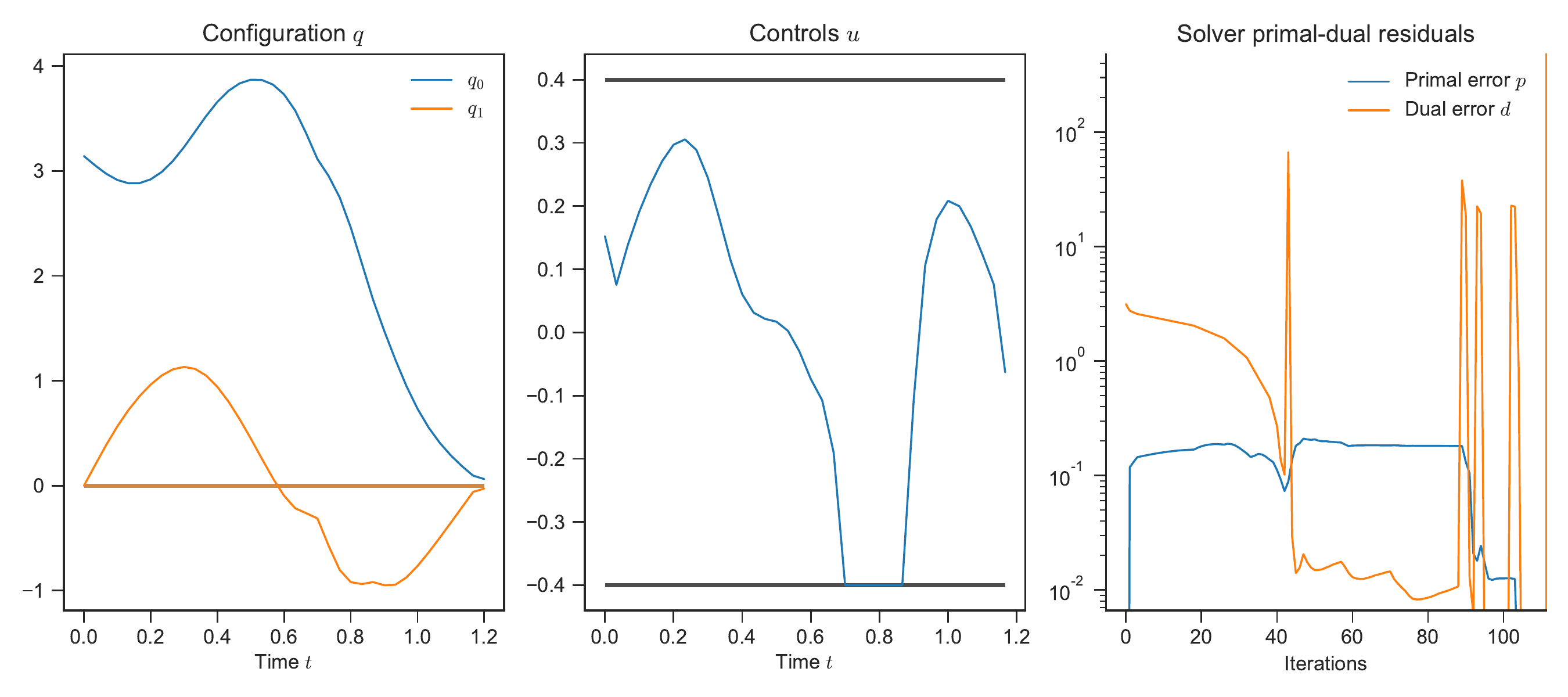}
	\caption{Angle and torque trajectory of the double-pendulum system, as well as the primal-dual convergence criteria. The controls saturate the imposed limit for a duration of \SI{200}{\milli\second}.}
	\label{fig:doublependulumtrajectories}
\end{figure}

We implemented a simple double-pendulum problem as an NLP using \proxnlp, Pinocchio and CasADi \cite{anderssonCasADiSoftwareFramework2019}, with a time step of $\Delta t = 30 \mathrm{ms}$ and desired convergence threshold of $\epsilon = 10^{-4}$. Here, the second-order derivatives of the dynamics are ignored in the Hessian computation. See fig.~\ref{fig:doublependulumtrajectories}.

\paragraph{Obstacle avoidance on UR10}

\begin{figure}[ht!]
	\vspace{2mm}
	\centering
	\def\snapleft{700pt}\def\snapright{560pt}
	\def\snapbot{200pt}
	\def\snaptop{220pt}%
	\def\snapleftB{560pt}\def\snaprightB{500pt}%
	\def\snapbotB{160pt}
	\def\snaptopB{60pt}
	\setlength{\tabcolsep}{0.1mm}
	\renewcommand{\arraystretch}{0.5}
	\def\snapwidth{.33\linewidth-\tabcolsep}
	\begin{tabular}{@{}ccc}
		\includegraphics[trim=\snapleftB{} \snapbotB{} \snaprightB{} \snaptopB{},clip=true,width=\snapwidth]{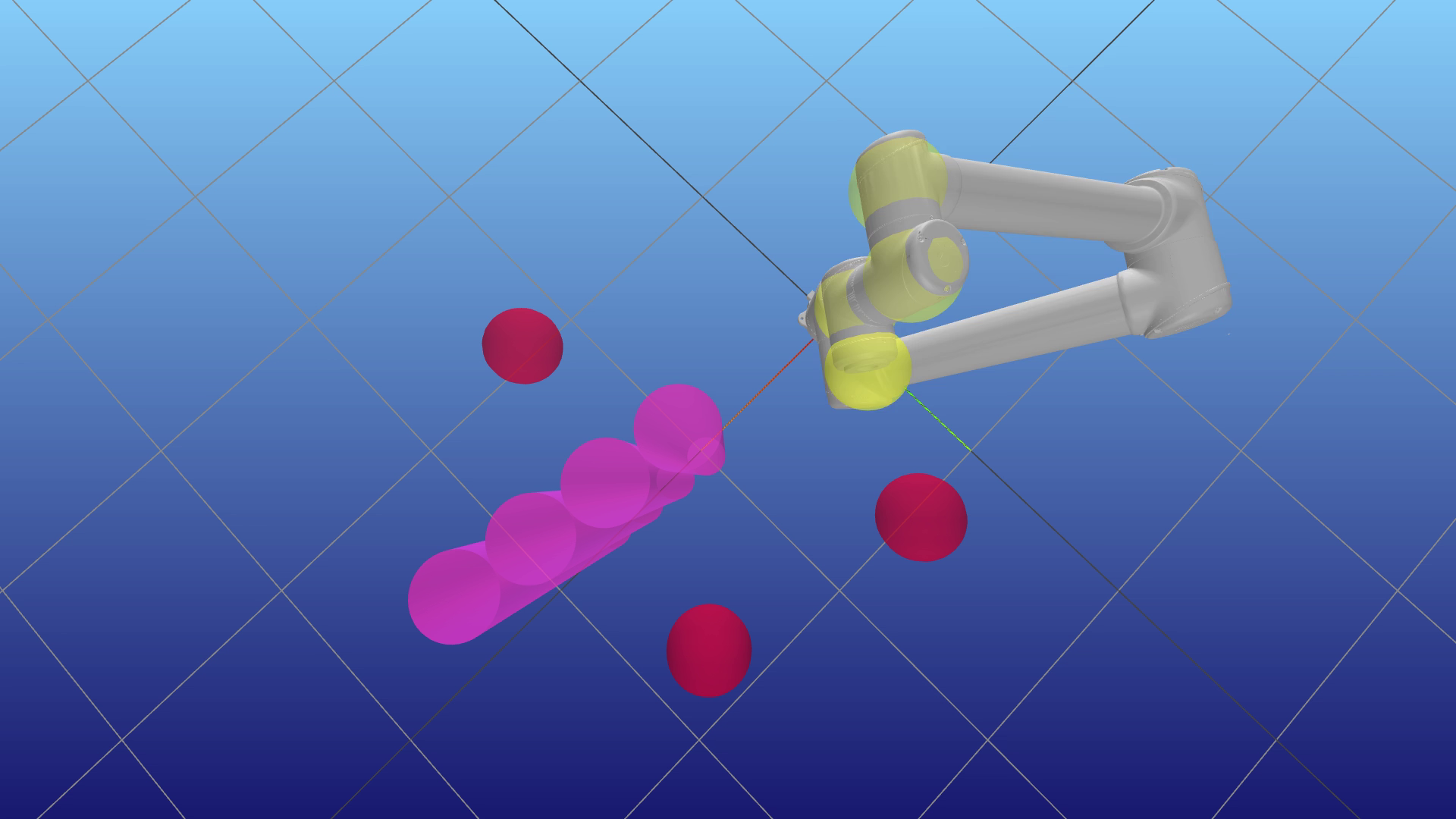} &
		\includegraphics[trim=\snapleftB{} \snapbotB{} \snaprightB{} \snaptopB{},clip=true,width=\snapwidth]{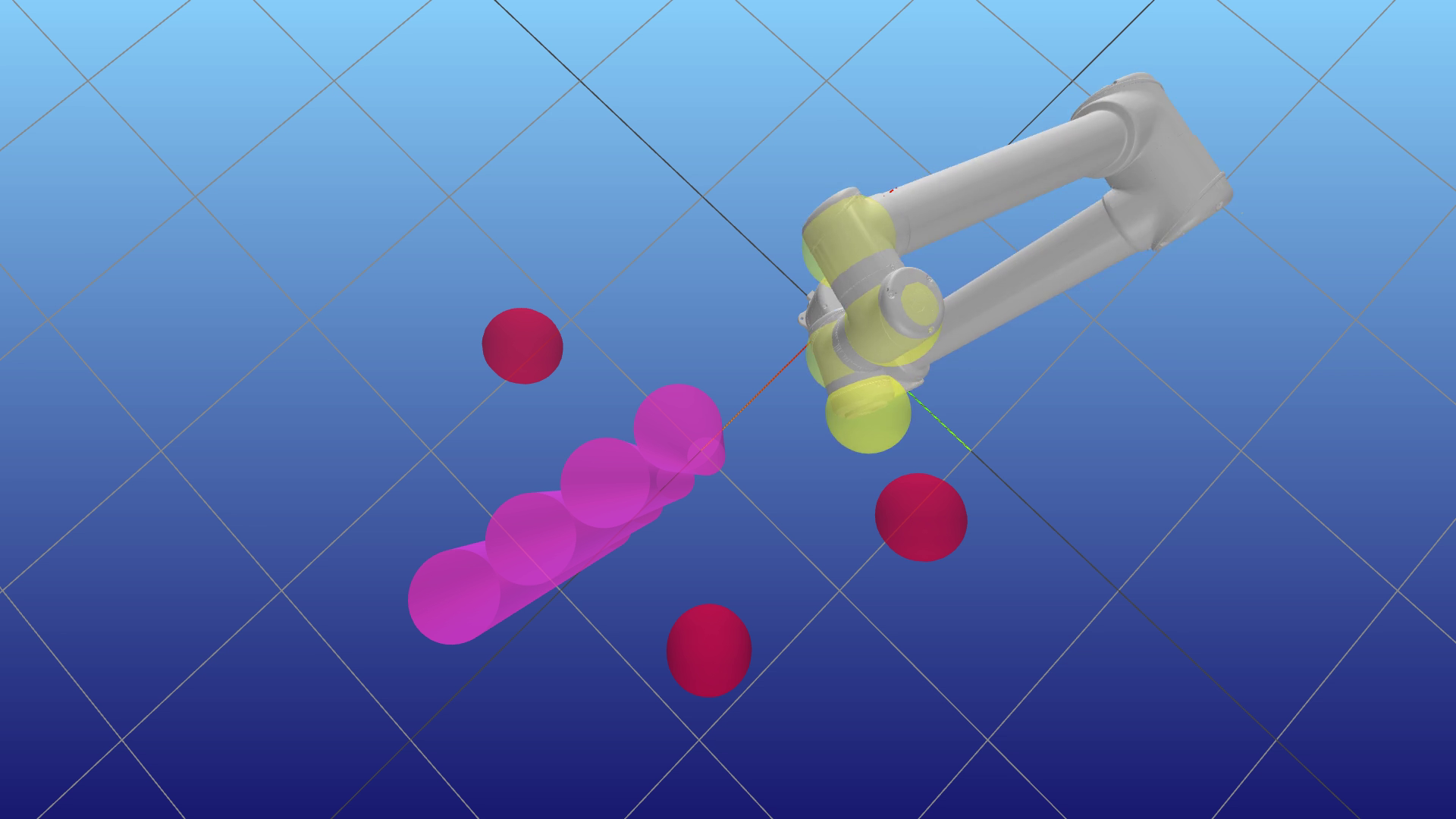} &
		\includegraphics[trim=\snapleftB{} \snapbotB{} \snaprightB{} \snaptopB{},clip=true,width=\snapwidth]{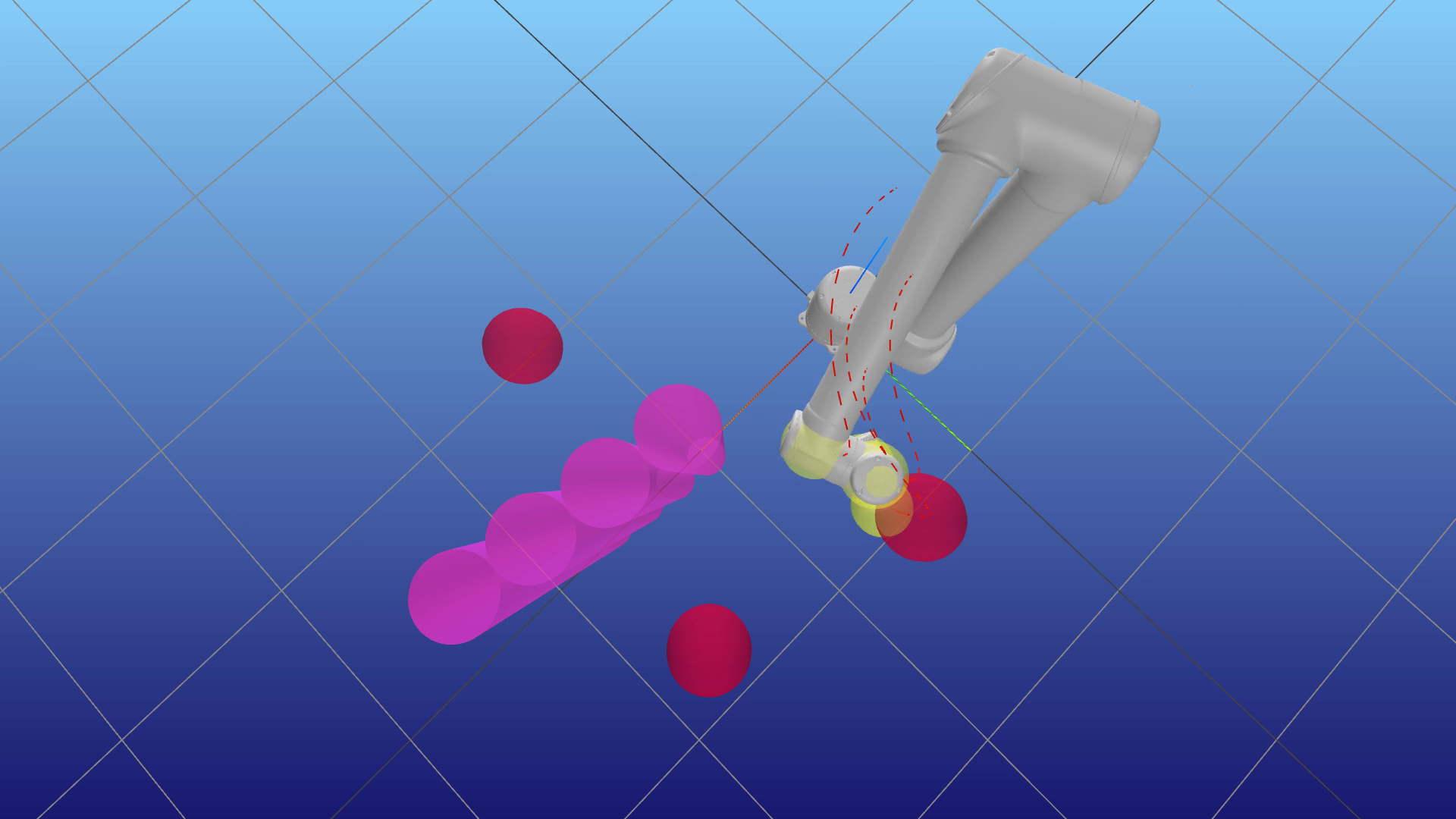} \\
		\includegraphics[trim=\snapleftB{} \snapbotB{} \snaprightB{} \snaptopB{},clip=true,width=\snapwidth]{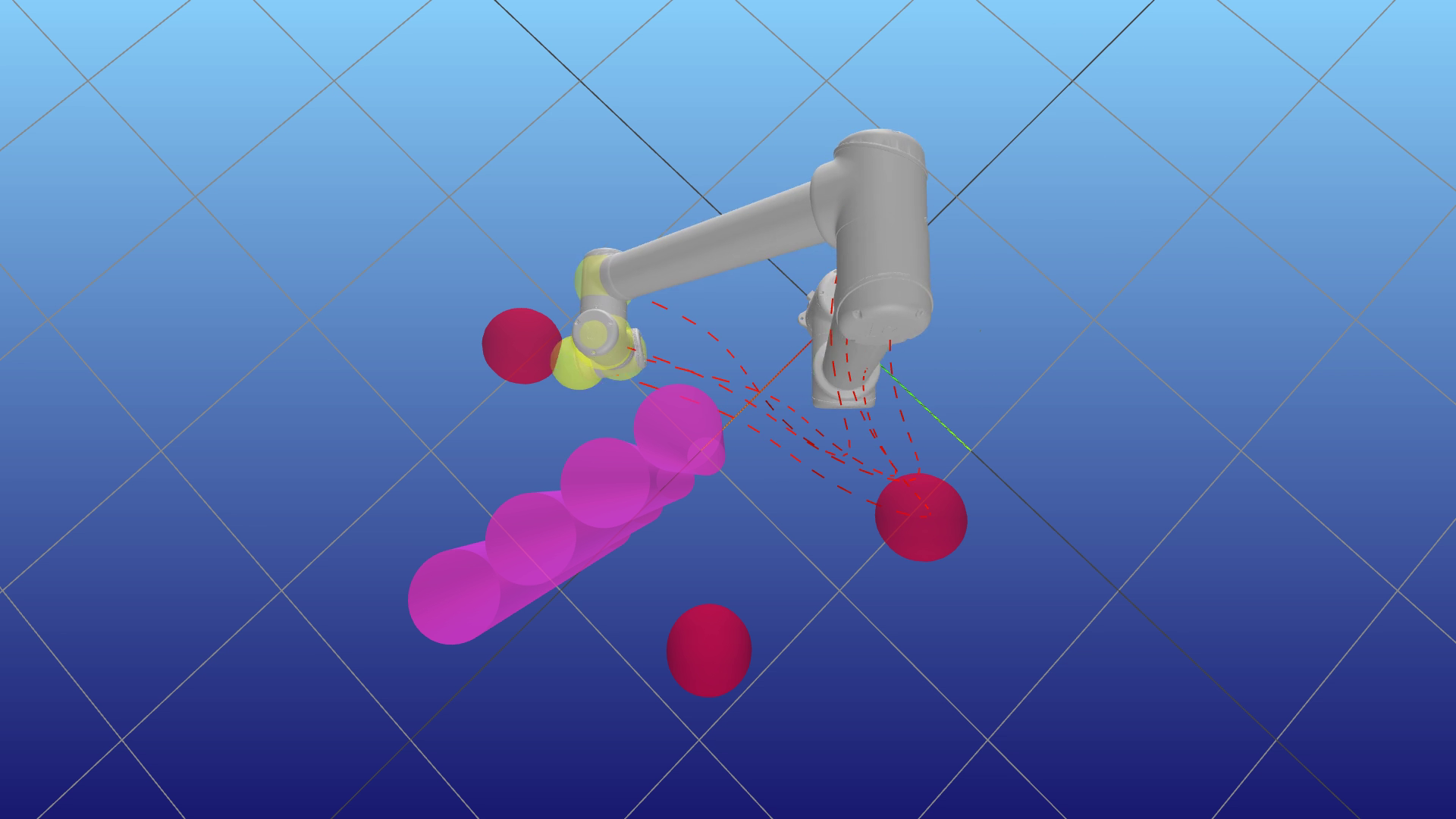} &
		\includegraphics[trim=\snapleftB{} \snapbotB{} \snaprightB{} \snaptopB{},clip=true,width=\snapwidth]{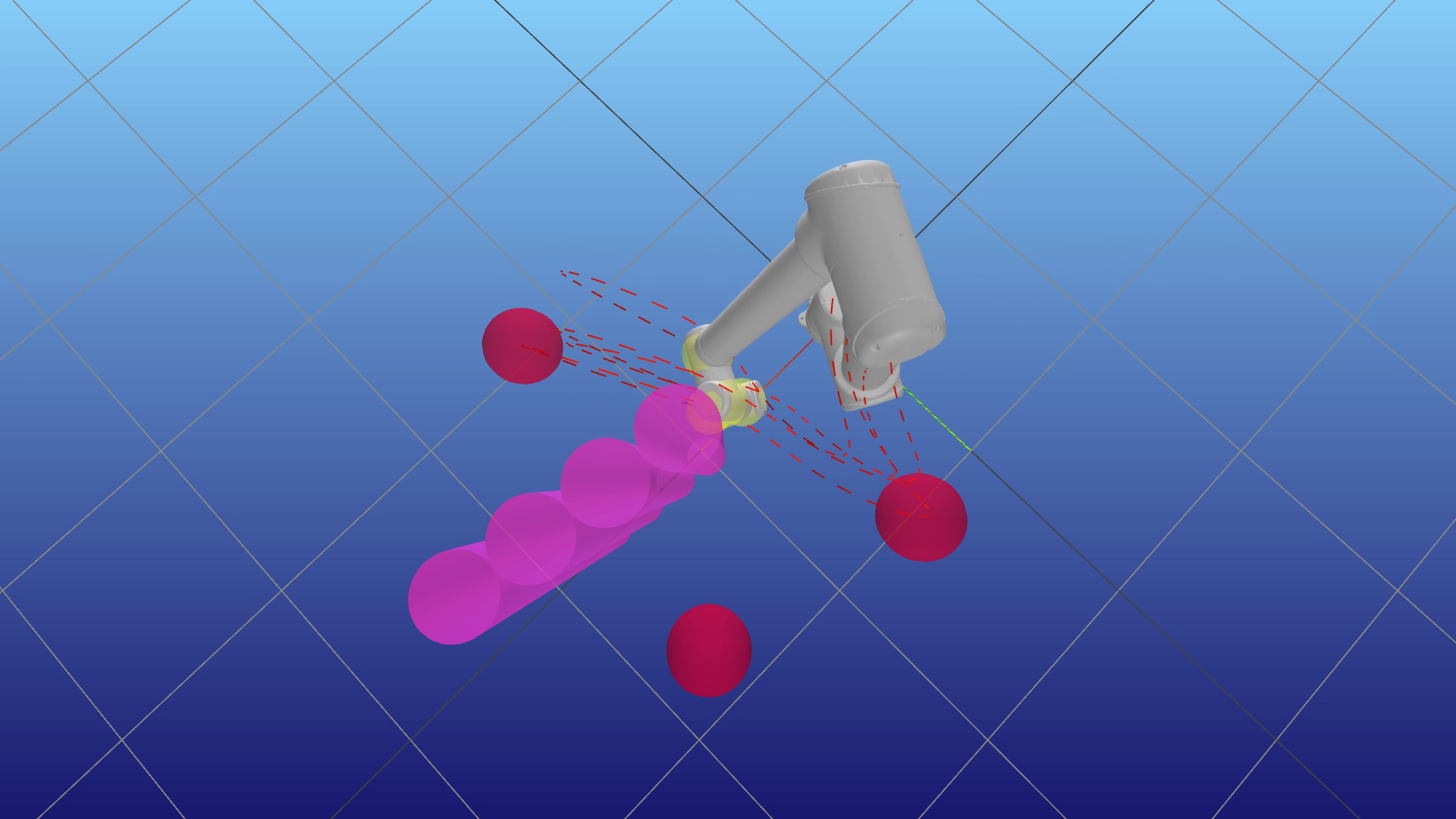} &
		\includegraphics[trim=\snapleftB{} \snapbotB{} \snaprightB{} \snaptopB{},clip=true,width=\snapwidth]{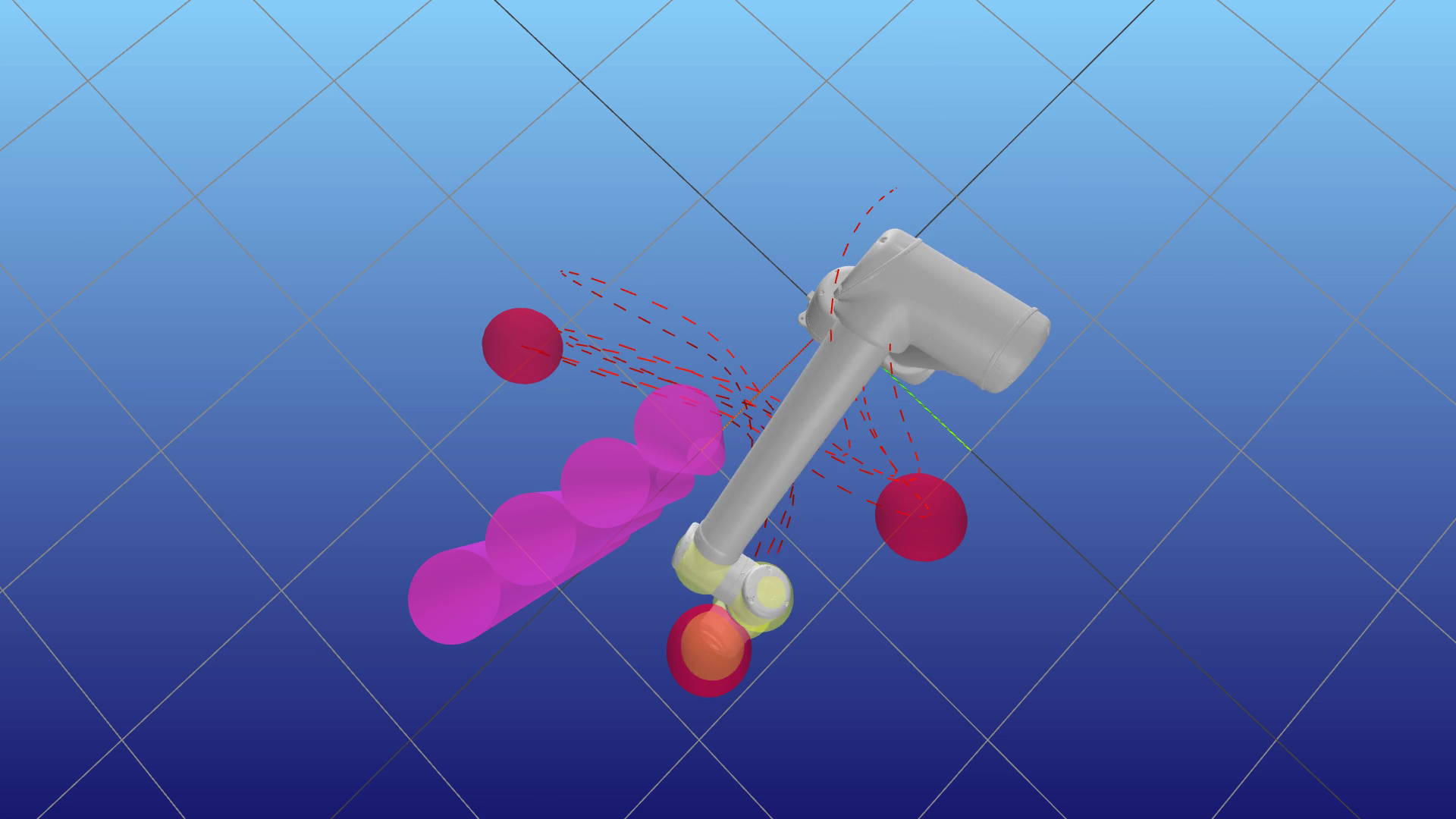}
	\end{tabular}
	\caption{UR10 reach task. The yellow spheres around the end-effector and wrist links do not collide with the purple cylinders, and the waypoints are reached at the specified times.}
	\label{fig:ur10_motions}
	\vspace*{-4mm}
\end{figure}

This example from our recent preprint~\cite{jalletConstrainedDifferentialDynamic2022} was implemented using our experimental code applying the method to DDP. See fig.~\ref{fig:ur10_motions}.

\begin{figure}[h]
	\centering
	\includegraphics[trim=680pt 0 680pt 200,clip=true,width=.96\linewidth]{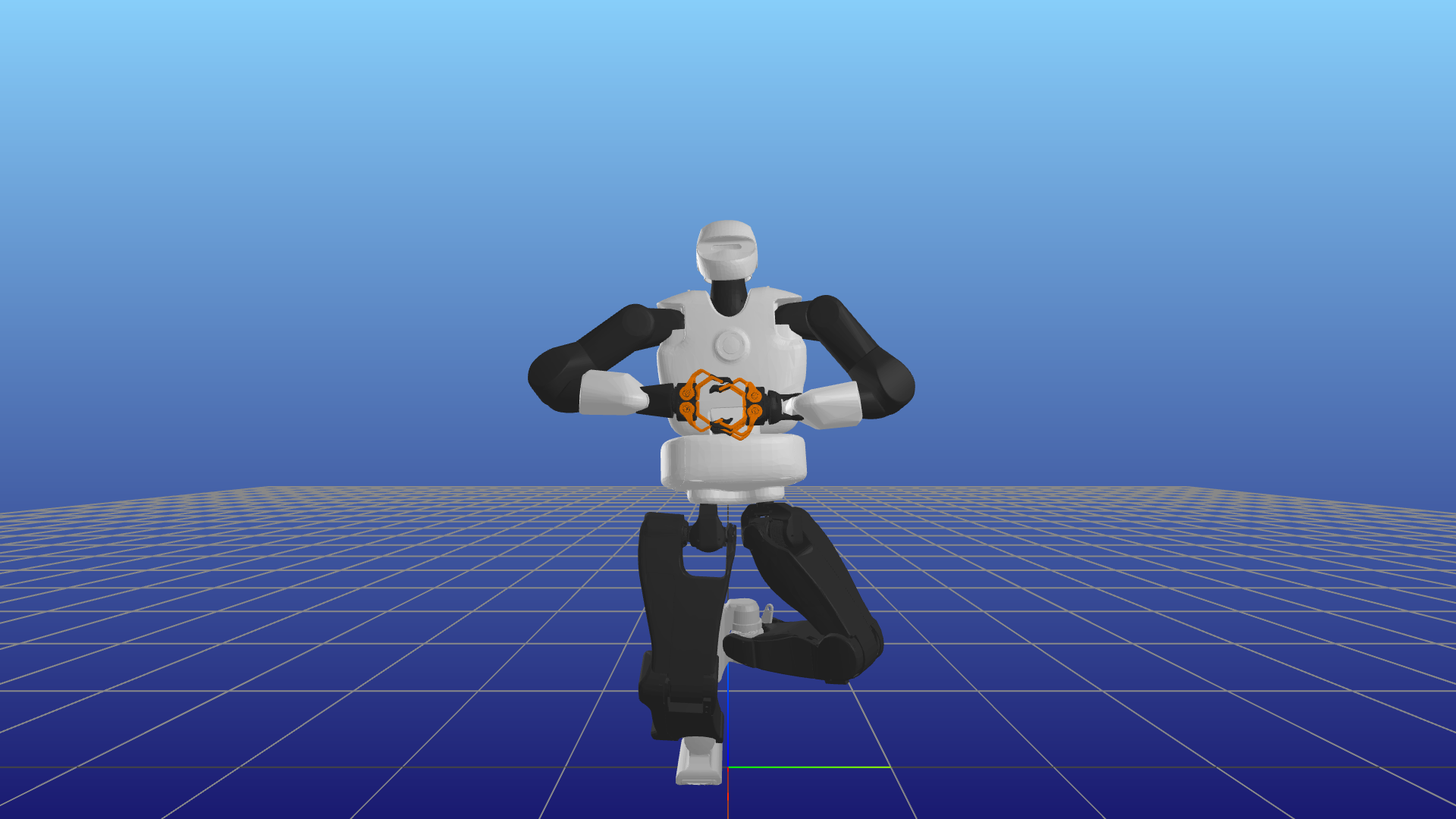}
	\caption{Generated pose on the Talos robot.}
	\label{fig:talos_yoga}
\end{figure}

\paragraph{Pose generation on Talos}
See figure \ref{fig:talos_yoga}.
The cost function reads, for configuration $q \in \calQ$
\begin{equation}
\begin{alignedat}{3}
	J(q) &=  0.1\|q\ominus q_0\|^2 \\&+ 2.5\| R_\mathrm{base}(q) \ominus R_\mathrm{base}(q_0) \|^2
	\\
	&+ 10\| p_{\mathrm{lg}}(q) - p_{\mathrm{rg}}(q) - \bfm{d} \|^2 & \text{(gripper dist.)} \\
	&+ 2 \| (p_\mathrm{le,y}(q),p_\mathrm{re,y}(q)) - (2, -2)\|^2 & \text{(elbow $y$)} \\
	&+ \|R_\mathrm{lg}(q) \ominus R_0\|^2 + \|R_\mathrm{rg}(q) \ominus R_1\|^2 & \text{(hand orn.)} \\
\end{alignedat}
\end{equation}
where $\bfm{d} = (0, 0.03, 0)^\top$, $\mathrm{lg, rg}$ mean left and right gripper, $\mathrm{le, re}$ mean left and right elbow, $R_\mathrm{base}$ is the body base orientation.
We have additional constraints: the right foot must be flat on the ground, the left foot must be $\geq \SI{40}{\centi\meter}$ above ground with $p_{\mathrm{lf},xy} \in [-0.05, 0.1]$ and a specific orientation, and the right gripper satisfies $p_{\mathrm{rg},y} \leq 0, p_{\mathrm{rg},z} \in [-1.1, 1.2]$.
The costs and constraints are implemented using CasADi~\cite{anderssonCasADiSoftwareFramework2019}.

\begin{figure}[ht!]
	\includegraphics[width=\linewidth]{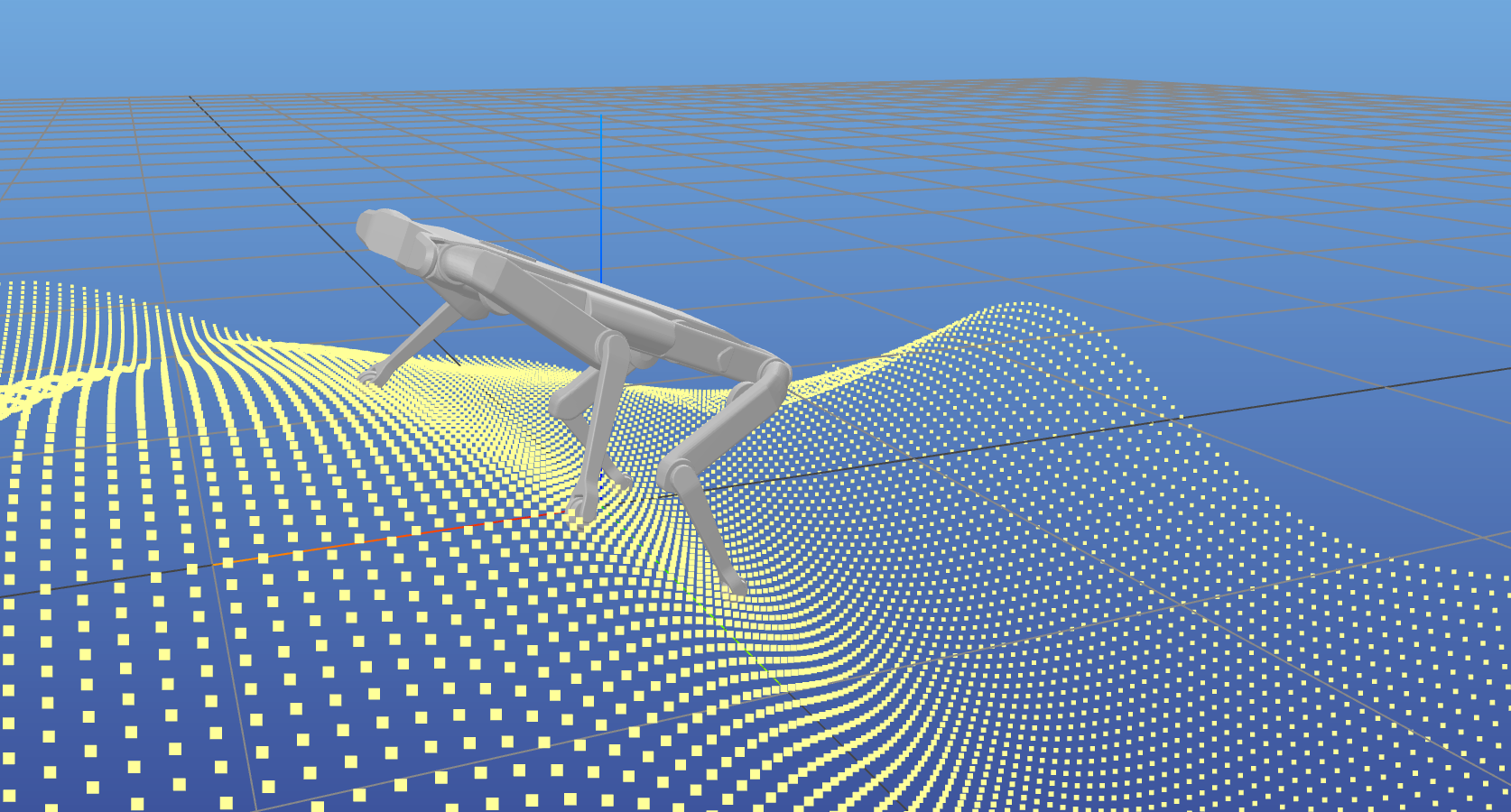}
	\caption{Inverse geometry Solo-12 quadruped with a heightmap and accounting for contact forces.}
	\label{fig:solo_heightmap}
\end{figure}

\paragraph{Solo inverse geometry with heightmap}
See figure \ref{fig:solo_heightmap}. The objective is to generate a feasible pose for the Solo-12 quadruped along with the 3D contact forces at the $4$ feet:
\begin{equation}
	\min_{q,\{f\}} \| \theta - \theta_0 \|^2 + \tfrac{1}{10}\sum_{i=1}^4 \|f_i\|^2,
\end{equation}
where $\theta,\theta_0$ are the joint angles of the robot (the pose without the base placement).
The problem has the following constraints:
\begin{itemize}
	\item zero angular momentum at the CoM
	\item the CoM altitude must be higher than the average foot altitude
	\item contact forces sum to the robot's weight
	\item contact forces satisfy the friction cone.
\end{itemize}

\section*{Acknowledgements}

We would like to thank Fabian Schramm and Guilhem Saurel for their help on testing and packaging the software, and Alessandro Assirelli for providing the pose generation examples on Talos and Solo.

This work was supported in part by the HPC resources from GENCI-IDRIS (Grant AD011011342), the French government under management of
Agence Nationale de la Recherche as part of the ”Investissements d’avenir”
program, reference ANR-19-P3IA-0001 (PRAIRIE 3IA Institute) and ANR-19-P3IA-000 (ANITI 3IA Institute), Louis Vuitton ENS Chair on Artificial
Intelligence, and the European project MEMMO (Grant 780684).

\balance
{
	\footnotesize
	\bibliographystyle{IEEEtran}
	\bibliography{references}

\begin{thebibliography}{10}
\providecommand{\url}[1]{#1}
\csname url@rmstyle\endcsname
\providecommand{\newblock}{\relax}
\providecommand{\bibinfo}[2]{#2}
\providecommand\BIBentrySTDinterwordspacing{\spaceskip=0pt\relax}
\providecommand\BIBentryALTinterwordstretchfactor{4}
\providecommand\BIBentryALTinterwordspacing{\spaceskip=\fontdimen2\font plus
\BIBentryALTinterwordstretchfactor\fontdimen3\font minus
  \fontdimen4\font\relax}
\providecommand\BIBforeignlanguage[2]{{%
\expandafter\ifx\csname l@#1\endcsname\relax
\typeout{** WARNING: IEEEtran.bst: No hyphenation pattern has been}%
\typeout{** loaded for the language `#1'. Using the pattern for}%
\typeout{** the default language instead.}%
\else
\language=\csname l@#1\endcsname
\fi
#2}}

\bibitem{jalletConstrainedDifferentialDynamic2022}
W.~Jallet, A.~Bambade, N.~Mansard, and J.~Carpentier, ``Constrained
  {{Differential Dynamic Programming}}: {{A}} primal-dual augmented
  {{Lagrangian}} approach,'' Mar. 2022.

\bibitem{murrayMathematicalIntroductionRobotic2017}
R.~M. Murray, Z.~Li, and S.~S. Sastry, \emph{A {{Mathematical Introduction}} to
  {{Robotic Manipulation}}}, 1st~ed.\hskip 1em plus 0.5em minus 0.4em\relax
  {CRC Press}, Dec. 2017.

\bibitem{hestenesMultiplierGradientMethods1969}
M.~R. Hestenes, ``Multiplier and gradient methods,'' \emph{Journal of
  Optimization Theory and Applications}, vol.~4, no.~5, Nov. 1969.

\bibitem{powellAlgorithmsNonlinearConstraints1978}
M.~J.~D. Powell, ``Algorithms for nonlinear constraints that use lagrangian
  functions,'' \emph{Mathematical Programming}, vol.~14, no.~1, Dec. 1978.

\bibitem{rockafellarMultiplierMethodHestenes1973}
R.~T. Rockafellar, ``The multiplier method of {{Hestenes}} and {{Powell}}
  applied to convex programming,'' \emph{Journal of Optimization Theory and
  Applications}, vol.~12, no.~6, Dec. 1973.

\bibitem{gillPrimaldualAugmentedLagrangian2012}
P.~E. Gill and D.~P. Robinson, ``A primal-dual augmented {{Lagrangian}},''
  \emph{Computational Optimization and Applications}, vol.~51, no.~1, Jan.
  2012.

\bibitem{howellALTROFastSolver2019}
T.~A. Howell, B.~E. Jackson, and Z.~Manchester, ``{{ALTRO}}: {{A Fast Solver}}
  for {{Constrained Trajectory Optimization}},'' in \emph{2019 {{IEEE}}/{{RSJ
  International Conference}} on {{Intelligent Robots}} and {{Systems}}
  ({{IROS}})}.\hskip 1em plus 0.5em minus 0.4em\relax {Macau, China}: {IEEE},
  Nov. 2019.

\bibitem{kazdadiEqualityConstrainedDifferential2021}
S.~Kazdadi, J.~Carpentier, and J.~Ponce, ``Equality {{Constrained Differential
  Dynamic Programming}},'' in \emph{{{ICRA}} 2021 - {{IEEE International
  Conference}} on {{Robotics}} and {{Automation}}}, May 2021.

\bibitem{jalletImplicitDifferentialDynamic2022}
W.~Jallet, N.~Mansard, and J.~Carpentier, ``Implicit {{Differential Dynamic
  Programming}},'' in \emph{International {{Conference}} on {{Robotics}} and
  {{Automation}} ({{ICRA}} 2022)}.\hskip 1em plus 0.5em minus 0.4em\relax
  {Philadelphia, United States}: {IEEE Robotics and Automation Society}, May
  2022.

\bibitem{bambadeProxQP}
A.~Bambade, S.~El-Kazdadi, A.~Taylor, and J.~Carpentier, ``Prox-qp: Yet another
  quadratic programming solver for robotics and beyond,'' in \emph{{Robotics:
  Science and Systems 2022}}, 2022.

\bibitem{rockafellarAugmentedLagrangiansApplications1976}
R.~T. Rockafellar, ``Augmented {{Lagrangians}} and {{Applications}} of the
  {{Proximal Point Algorithm}} in {{Convex Programming}},'' \emph{Mathematics
  of Operations Research}, vol.~1, no.~2, 1976.

\bibitem{eigenweb}
G.~Guennebaud, B.~Jacob, \emph{et~al.}, ``Eigen v3,''
  http://eigen.tuxfamily.org, 2010.

\bibitem{carpentierPinocchioLibraryFast2019}
J.~Carpentier, G.~Saurel, G.~Buondonno, J.~Mirabel, F.~Lamiraux, O.~Stasse, and
  N.~Mansard, ``The {{Pinocchio C}}++ library \textendash{} {{A}} fast and
  flexible implementation of rigid body dynamics algorithms and their
  analytical derivatives,'' in \emph{{{IEEE International Symposium}} on
  {{System Integrations}} ({{SII}})}, 2019.

\bibitem{carpentierAnalyticalDerivativesRigid2018}
J.~Carpentier and N.~Mansard, ``Analytical {{Derivatives}} of {{Rigid Body
  Dynamics Algorithms}},'' in \emph{Robotics: {{Science}} and {{Systems
  XIV}}}.\hskip 1em plus 0.5em minus 0.4em\relax {Robotics: Science and Systems
  Foundation}, June 2018.

\bibitem{anderssonCasADiSoftwareFramework2019}
J.~A.~E. Andersson, J.~Gillis, G.~Horn, J.~B. Rawlings, and M.~Diehl,
  ``{{CasADi}}: A software framework for nonlinear optimization and optimal
  control,'' \emph{Mathematical Programming Computation}, vol.~11, no.~1, Mar.
  2019.

\end{thebibliography}
}

\end{document}